\documentclass{ecai}
\usepackage{times}
\usepackage{graphicx}
\usepackage{latexsym}
\usepackage{amsmath,amssymb,amsfonts}
\usepackage{caption}
\usepackage[misc]{ifsym}

\begin{document}

\title{Dedge-AGMNet:an effective stereo matching network\\
optimized by depth edge auxiliary task}

\author{Weida Yang\institute{ The Key Laboratory of Integrated Microsystems, Shenzhen Graduate School, Peking University, China. email:weida.yang@pku.edu.cn, yongzhao@pkusz.edu.cn, *corresponding author} \and  Xindong Ai$^1$ \and Zuliu Yang$^1$ \and Yong Xu\institute{Harbin Insititute of Technology} \and  Yong Zhao$^{1*}$}
\maketitle
\bibliographystyle{ecai}

\begin{abstract}
To improve the performance in ill-posed regions, this paper proposes an atrous granular multi-scale network based on depth edge subnetwork(Dedge-AGMNet). According to a general fact, the depth edge is the binary semantic edge of instance-sensitive. This paper innovatively generates the depth edge ground-truth by mining the semantic and instance dataset simultaneously. To incorporate the depth edge cues efficiently, our network employs the hard parameter sharing mechanism for the stereo matching branch and depth edge branch. The network modifies SPP to Dedge-SPP, which fuses the depth edge features to the disparity estimation network. The granular convolution is extracted and extends to 3D architecture. Then we design the AGM module to build a more suitable structure. This module could capture the multi-scale receptive field with fewer parameters. Integrating the ranks of different stereo datasets, our network outperforms other stereo matching networks and advances state-of-the-art performances on the Sceneflow, KITTI 2012 and KITTI 2015 benchmark datasets.
\end{abstract}

\section{INTRODUCTION}
Visual perception is a fundamental problem that focuses on the capability to obtain accurate results in a 3D scene. Depth estimation is an important part of perception, which has various essential applications, such as autonomous driving, dense reconstruction, and robot navigation. As a type of passive depth sensing techniques, stereo matching estimates the disparity from rectified image pairs.

The classical pipeline for disparity estimation involves finding corresponding points based on matching cost and post-processing. With the development of deep learning, learning-based methods acquire cues from classical ones, they are embedded in different modules that attempt to obtain a better result. However, because of the discontinuous inference process and the shallow features, the early CNN-based methods capture a terrible performance in ill-posed regions. Nowa-days, the end-to-end disparity estimation network is proposed to improve the performance.

Currently, there are two main methods to optimize the networks in ill-posed regions. The first approach captures the additional features and constraints using auxiliary networks, such as semantic segmentation and edge detection subnetworks\cite{du2019amnet,song2019edgestereo,yang2018segstereo}. However, the semantic segmentation tasks are incapable to distinct the overlap instance with the same label. And the classical edge information contains a large number of noise edges. Those issues induce disparity estimation misjudgments. Secondly, some network utilize a set of stacked 3D convolution modules\cite{chang2018pyramid,du2019amnet} or parallel structures \cite{chabra2019stereodrnet} to capture multi-scale context information. These methods are useful but greatly increase computational consumption and memory resources.

In view of the above problems, this paper proposes a multi-task learning network called Dedge-AGMNet that effectively alleviates the drawbacks of both previous methods. We  generate depth edge ground-truth and propose the depth edge auxiliary network. Sharing the feature extraction module with the stereo matching main network, the auxiliary branch provides the depth edge constraints. For effective multi-task interactions, we design the Dedge-SPP that embeds the depth edge features to the main branch. Compared with traditional edge detection, the proposed network substantially reducing the noise edges.

The paper proposes a novel module, called the AGM module. Referring to Res2Net\cite{gao2019res2net}, we extract the granular convolution from its block and extend to the 3D representation. Retaining the advantages of multi and large scale receptive field, we employ the parallel structure to trade-off the running latency and the scale of the receptive field. The main contributions of this work are summarized as follows:
\begin{figure}[t]
	\centering
	\includegraphics[width=0.95\columnwidth]{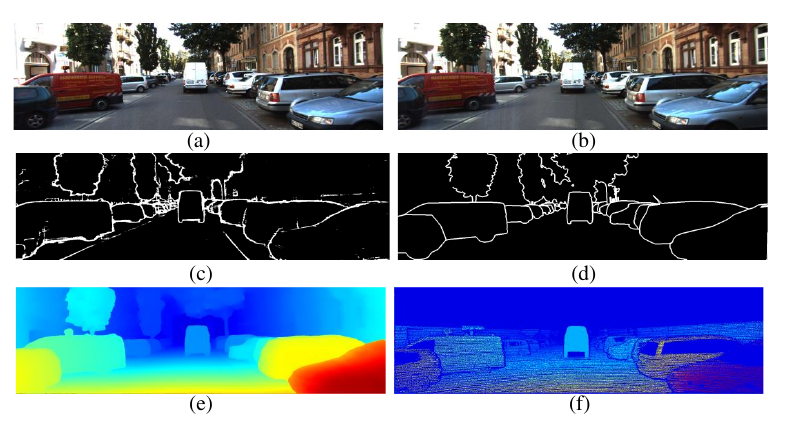}  
	\caption{(a)\&(b) the left and right images from KITTI 2015; (c)\&(e) the predicted depth edge and disparity map from training set; (d)\&(f) the ground-truth of the depth edge and the disparity estimation.}
	\label{fig1}
\end{figure}

\begin{figure*}[t]
	\centering
	\includegraphics[width=0.93\textwidth]{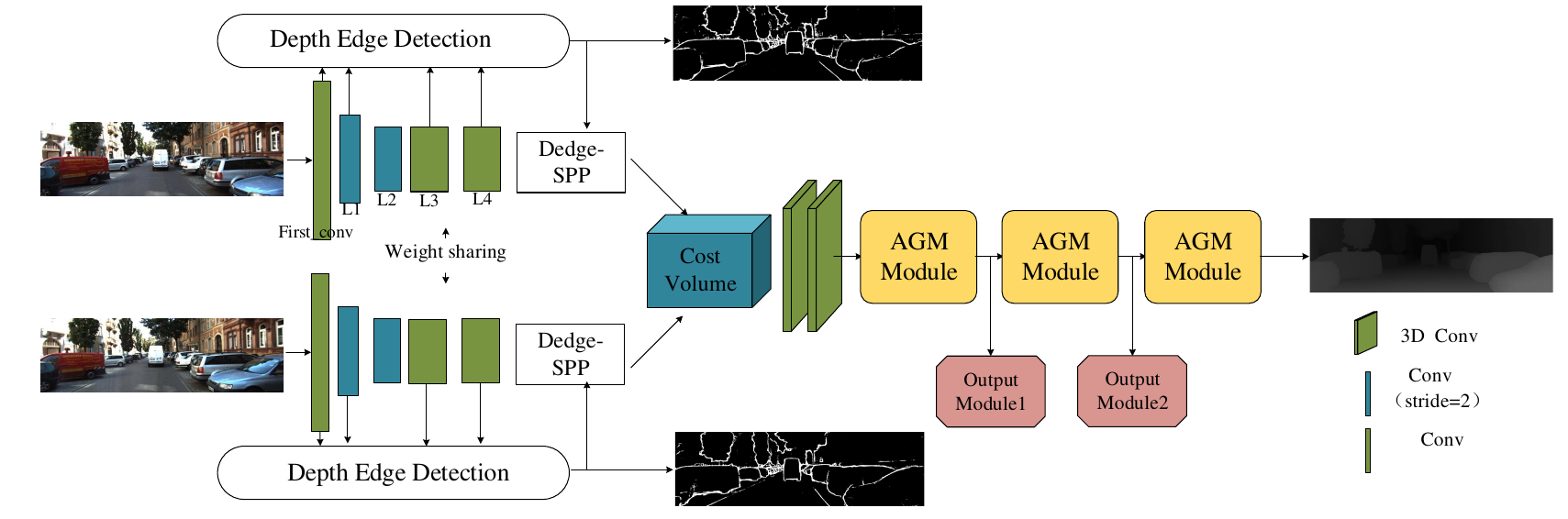} 
	\caption{The pipeline of the proposed atrous granular multi-scale network based on depth edge subnetwork(Dedge-AGMNet).}
	\label{fig2}
\end{figure*}
\begin{itemize}
	\item We propose the multi-task learning network Dedge-AGMNet that optimizes the feature extraction module with hard sharing parameter, and utilizes the Dedge-SPP to incorporate depth edge cues into disparity estimation pipeline.
	
	\item The AGM module is designed to capture the multi-scale information while requiring fewer parameters at a reduced computational cost.
	
	\item Our method achieves state-of-the-art accuracy on the Sceneflow dataset, KITTI 2012 and KITTI 2015 stereo benchmark.	
	
\end{itemize}

\section{RELATED WORK}
\subsection{Stereo Matching}
Depth from stereo has been widely studied for a long time in the literature. The traditional stereo matching methods\cite{schops2017multi} have been proposed for four steps: matching cost computation\cite{zbontar2016stereo}, cost aggregation\cite{luo2016efficient}, optimization\cite{seki2017sgm}, and disparity refinement. Recently, convolutional neural networks have become popular in solving this problem. Zbontar and LeCun \cite{zbontar2016stereo} were the first to use CNN for matching cost computation. Luo et al.\cite{luo2016efficient} designed a novel Siamese network to treat the computation of matching cost as a multi-label classification, which computes the inner product between the left and the right feature maps. Seki et al.\cite{seki2017sgm} raised the SGM-Net that predicts SGM penalties for regularization.

Inspired by other pixel-wise labeling tasks, the end-to-end neural networks have been proposed using the fully-convolution network\cite{long2015fully} for disparity estimation. Mayer et al.\cite{mayer2016large} designed the first end-to-end disparity estimation network, DispNet, which utilizes the encoder-decoder structure with short-cut connections for second stage processing. Kendall et al.\cite{kendall2017end} raised GCNet, a cost volume formed by concatenating the feature maps to incorporate contextual information. This network applies the 3D encoder-decoder architecture to regularize the cost volume. To find correspondences in ill-posed regions, Chang and Chen\cite{chang2018pyramid} proposed the PSMNet to regularize cost volume using stacked multiple hourglass networks in conjunction with intermediate supervision.

Currently, Chabra\cite{chabra2019stereodrnet} proposed a depth refinement architecture that helps the fusion system to produce geometrically consistent reconstructions, and utilized 3D dilated convolutions to construct hourglass architecture. Meanwhile, XianZhi Du et al.\cite{du2019amnet} designed a similar construction with three atrous multi-scale modules, while it is useful to aggregate rich multi-scale contextual information from cost volume. Base on the PSMNet \cite{chang2018pyramid}, both of them achieved state of the art on the different stereo datasets.

\subsection{Multi-scale Features}
The multi-scale feature is an important factor in the pixel predicted tasks, such as semantic segmentation and disparity estimation. Because ambiguous pixels require a diverse range of contextual information, ASPP\cite{chen2017rethinking} was designed to concatenate various feature maps with multi-scale receptive fields. To further quest the importance of the receptive field, Yang et al.\cite{yang2018denseaspp} proposed Dense ASPP to concatenate a set of different atrous convolutional layers densely. The approach encourages feature reusing by constructing a similar structure with the DenseNet\cite{huang2017densely}. Instead of representing the multi-scale features in a layer-wise manner, Gao et al.\cite{gao2019res2net} designed novel architecture, called Res2Net. The network uses hierarchical residual-like connections in a single block to represent it at a granular level. Controlling the same computational resources as ResNet block, Res2Net achieved more accurate results.

\subsection{Multi-task Learning network}
Focus on improving the accuracy of the ill-posed regions where the single stereo matching networks are difficult to predict. Yang et al.\cite{yang2018segstereo} proposed SegStereo to embed the semantic features, and regularized semantic cues as the loss term. Xianzhi Du et al.\cite{du2019amnet} utilized foreground-background segmentation map to improve disparity estimation. This paper believed that better awareness of foreground objects would lead to a more accurate estimation. Song et al.\cite{song2018edgestereo,song2019edgestereo} proposed EdgeStereo which composes a disparity estimation subnetwork and an edge detection subnetwork. By combining the advantages of the semantic segmentation and edge detection, we propose the depth edge detection auxiliary network.

\section{Dedge-AGMNet}

\begin{figure}[t]
	\centering
	\includegraphics[width=0.85\columnwidth]{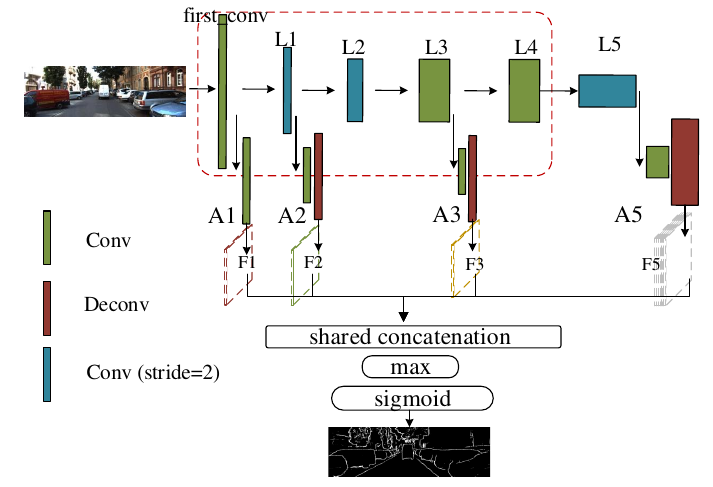}  
	\caption{The architecture of depth edge detection branch. The red dashed box denotes the sharing feature extraction module.}
	\label{fig3}
\end{figure}

The proposed Dedge-AGMNet is composed of a depth edge detection branch and a disparity estimation branch. The depth edge subnetwork provides geometric knowledge and constraints without adding irrelevant edges. We also utilize the granular convolution to design a more efficient 3D aggregate filtering module. It is worth noting that our network only estimates the disparity map but not predicts depth edge in the inference process, which decreases the parameters significantly.

\subsection{Network Architecture}
The structure of the proposed Dedge-AGMNet is shown in Fig.\ref{fig2}. The network consists of five parts, feature extraction, depth edge prediction and embedding, cost volume construction, 3D aggregation, and disparity prediction.

For the feature extraction from both subnetworks, we retain the ResNet-like structure used in PSMNet\cite{chang2018pyramid} except that the first downsampling operation occurs at L1 but not first\_conv.

We present the depth edge subnetwork with corresponding loss function to provide geometrical constraints for the shared features. In addition, instead of SPP\cite{chang2018pyramid}, the Dedge-SPP module is constructed to fuses the geometric knowledge from depth edge subnetwork. The details are described in Section 3.2.

The cost volume consists of two parts: a concatenation volume and a distance volume, which is explained in Section 3.4. We process the cost volume with a pre-hourglass module and three stacked AGM modules. And details are described in Section 3.3.

In the disparity estimation subnetwork, the three stacked AGM modules are connected to output modules to predict disparity maps. The details of the output modules and the loss functions are described in Section 3.4.

\subsection{Depth edge auxiliary task}
\subsubsection{Validity analysis \& Generation of dataset}
\begin{figure}[t]
	\centering
	\includegraphics[width=0.95\columnwidth]{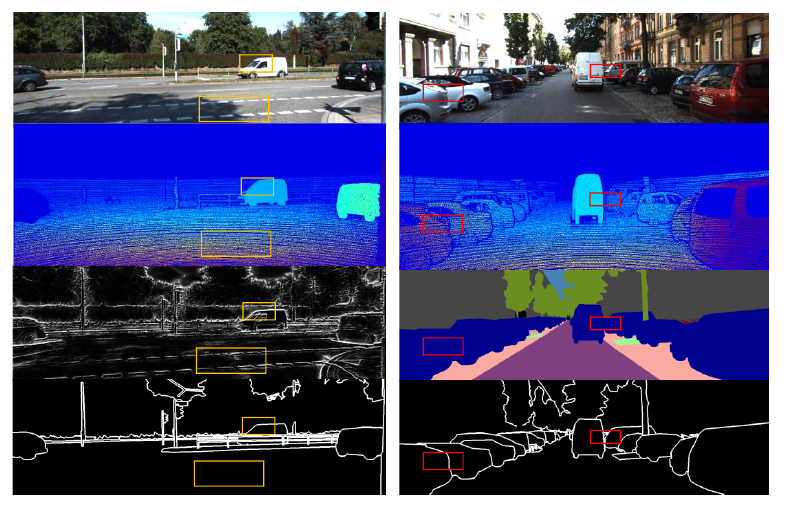}  
	\caption{The first row denotes left images, the second row presents the disparity ground-truth, the third row shows the classical edge and the semantic segmentation map, and the last row displays the depth edge maps. The yellow box presents the noise edge performance in the smooth region. The red box compares the performance in the disparity change region.}
	\label{fig4}
\end{figure}

Without additional knowledge or constraints, it is difficult to find correct correspondences in ill-posed regions. The classical edge subnetwork\cite{song2018edgestereo,song2019edgestereo} is beneficial, but it captures considerable edge noises, such as object pattern and inner edges. This non-semantic information heavily interferes with the disparity estimation. Semantic segmentation subnetworks\cite{yang2018segstereo,du2019amnet} are commonly used. However, semantic boundaries always lack the edge for overlap instances that have the same label, it induces disparity estimation misjudgment. As shown in Fig.\ref{fig4}, depth edge combines the advantages of classical edge and semantic map, it segments different individuals accurately without edge noise.

In the autonomous driving scene, a single foreground object always could be considered the same depth, this paper utilizes the binary instance bounds to represent the depth edges for the foreground object. We employ binary semantic boundaries to compensate for the lacking background edges. In summary, we generate the depth edge map by mining the instance\&semantic ground-truth in stereo datasets.

\subsubsection{Structure of subnetwork}

As shown in Fig.\ref{fig3}, except to share parameters in the feature extraction module, the auxiliary network adds L5(a similar structure with L4\cite{chang2018pyramid}) to capture more features. Different from the classical edge detection network\cite{xie2015holistically}, our network does not predict the depth edge based on bottom side features. But those features are useful to provide detailed edge information, we utilize the shared concatenation\cite{yu2017casenet} to fuse multi-frequency features, and only predict the edge at the last stage.

The bottom features $F = \{F_1, F_2, F_3\}$ are output from the feature re-extraction module$(A1, A2, A3)$, and the top features with K channels are represented by $F_5$. The shared concatenation is as follows:
$$\{F_5(1),F,F_5(2),F,...,F_5(K),F\}$$
the depth edge branch adopts a similar architecture as CASENet but contains several key modifications.

\begin{itemize}
	\item  The different task. Fist, compared with the semantic edge, the depth edge ground-truth consists of the boundaries from the instance and semantic map. It contains more useful edge information. Besides, we simplify the task from multi-label to binary label, which decreases the task complexity substantially. And drive the auxiliary branch to pay more attention to the edge details but not the classiﬁcation.
	\item Fewer parameters. Since the limitation of the parameters and com-putational cost, in contract to CASENet\cite{yu2017casenet}, we adopt about 1/8 channels in the feature extraction module. However, we believe that the simpliﬁed task could utilize fewer channels to capture the required features.
	\item Similar to CASENet, Our network handles more channels(K) for top features. But instead of building the relationship between each channel and corresponding label, we believe that more channels mean greater importance. Compared to the different probability from corresponding channels, the subnetwork selects the highest probability as the predicted probability for depth edge.
\end{itemize}

Besides, since the classical edge lacks semantic information, EdgeStereo\cite{song2019edgestereo} only shares parameters in shallow layers to capture low-frequency features. In constrast, The depth edge contains the semantic and instance information, our network still shares parameters in the high layers. The network could employ semantic information to suppress the interference of non-depth noise edges. We will illustrate it in Section 4.3.

\subsubsection{The incorporation of the networks}
This paper designs the depth edge loss function to optimize the sharing feature extraction. Because the depth edge label is a binary representation, we use the binary cross-entropy loss instead of multi-label\cite{yu2017casenet}. It denoted as $L_{edge}P(X_i; W)$ and $Y_i$ denote the predicted probability and ground-truth for the image pixel $X_i$.
\begin{equation}\label{eq1}
L_{edge}\left( X_i,W \right)=\left\{
\begin{array}{lr}
\alpha \times \log(1-P(X_i;W)), \: \: \: \ if\, Y_i = 0\\
\beta \times \log P(X_i;W),     \qquad \; \; \; \;\;  if\, Y_i = 1 &
\end{array}
\right .
\end{equation}
in which 
\begin{equation}\label{eq2}
\begin{array}{lr}
\alpha = \frac{\left|Y^+\right|}{\left|Y^+\right|+\left|Y^-\right|} \\
\beta = \frac{\left|Y^-\right|}{\left|Y^+\right|+\left|Y^-\right|}
\end{array}
\end{equation}
where $|Y^+|$ and $|Y^-|$ represent the number of positive samples and negative samples, respectively.

Since the disparity discontinuity point is always on the depth boundaries. the depth edge gradient is more consistent with the change of the disparity map, and $L_{dedge-disp}$ is presented as follows:
\begin{equation}\label{eq3}
L_{dedge-disp} = \frac{1}{N}\sum_{i, j}\left|\partial_xd_{i,j}\right|e^{-\gamma\left|\partial_x \xi_{i,j}\right|} + \left|\partial_yd_{i,j}\right|e^{-\gamma\left|\partial_y \xi_{i,j}\right|}
\end{equation}
where $N$ denotes the number of pixels, $\gamma$ is the loss intensity,  $\partial d$ and $\partial \xi$ present the disparity and the depth edge map gradient, respectively. 

What’s more, we concatenate the depth edge features with the output of L4 to modify the SPP\cite{chang2018pyramid}. Dedge-SPP is designed to share the geometric knowledge with the disparity estimation branch.

\subsection{Atrous granular multi-scale module}
\subsubsection{Structure of AGM module}
We propose AGM-module. As shown in Fig.\ref{fig5}, the AGM module combines the advantages of the hourglass and the parallel structure. The hourglass structure could reduce the feature size reasonably, we utilize the short-cut connection to transmit shallow features. Besides, the parallel structure with the dilated granular convolution boosts the performance significantly. Compared to the standard convolution, granular convolution captures multi-scale context information that requires fewer parameters. Meanwhile, the parallel structure balances the running latency and the scale of receptive field.

\subsubsection{Granular convolution}
\begin{figure}[t]
	\centering
	\includegraphics[width=0.95\columnwidth]{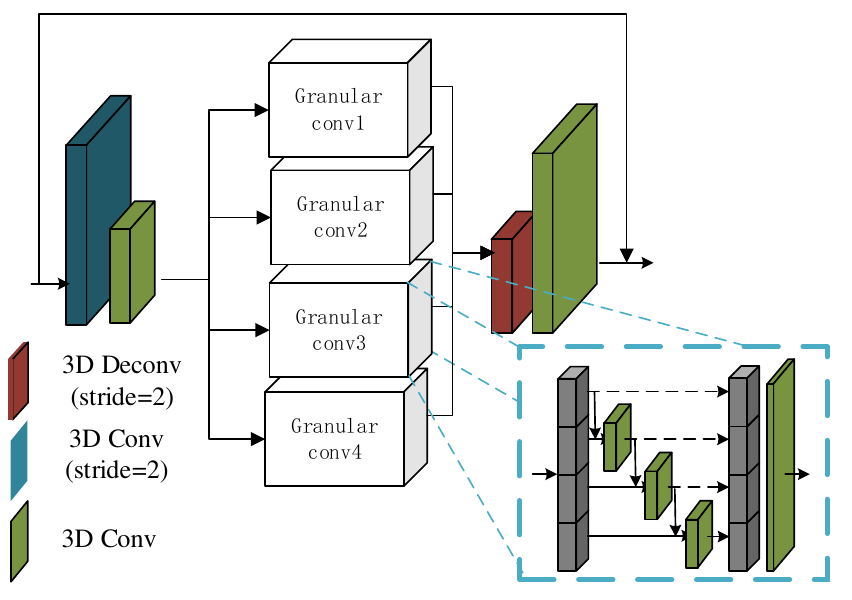}  
	\caption{The architecture of AGM module. The blue dashed box illustrates the granular convolution.}
	\label{fig5}
\end{figure}

The blue dashed box of Fig.\ref{fig5} illustrates the details of granular convolution. It shows that the number of receptive fields in granular convolution is approximate $G$ times than the standard convolution. The granular convolution divides the input features into several groups, the output features of the previous group are input to the next group of filters along with another group of input feature maps. The features map $v = (v_1, v_2, ..., v_G),  v_i \in R^{W \times H \times c/G(group number)}$. $\underline{\sum}$ denotes the concatenation operator and $<,>$ denotes standard convolution. We formulate granular convolution as follows:
\begin{equation}\label{eq4}
\begin{aligned}
&&	v' &= w_{pw} \underline{\sum}_{g=1}^G \hat{v}_g' \\
&&	   &= w_{pw} \underline{\sum}_{g=1}^G\sum_{i=1}^g <w_1 ...<w_i v_{(g-i)}>>
\end{aligned}
\end{equation}

\noindent where the weight $w=(w_1, w_2,...,w_G)$,  $w_i \in R^{\frac{c}{G} \times \frac{c}{G}  \times s \times s}$. And the $w_{pw}$ denotes the weight of point-wise convolution weight.

Keeping the channel and size of the feature map, the parameters of standard and granular convolution is shown below:

\noindent Standard convolution:
$$N_{standard} = C_{out} \times C_{in} \times s \times s = C^2 \times s^2$$
\noindent Granular convolution:
\begin{equation}\label{eq5}
\begin{aligned}
&&	N_{granular}  &= C_{in} \times C_{out} \times \frac{s}{C} \times \frac{s}{C} \times (G - 1) + C_{out} \times C_{out}  \\
&&	   		     &= C_{in} \times C_{out} \times s \times s \times (\frac{G - 1}{G \times G} + \frac{1}{s \times s}) \\
&&			     &\approx \frac{1}{G} \times N_{standard}
\end{aligned}
\end{equation}

\subsubsection{Running latency}
Granular convolution utilizes the internal cascade structure to capture the multi and large receptive field. However, As one layer is splitted into two or more sequential layers, the latency increases progressively. Therefore, this structure increases running latency inevitably. $K$ and $G$ denote the number of sequential layers and groups, respectively. The running latency of the cascade and parallel structures are shown as follow:
$$RL_{parallel} = G - 1 = 1/K \times R_{cascade}$$

In summary, contrasted with standard convolution, the computational cost of granular convolution is about $1/G$ times. The hyper-parameter $G=K=4$, AGM module utilizes the parallel structure to trade-off the running latency and the scale of the receptive field.

\subsection{Cost volume}
We designed the cost volume by stacking the concatenation module and the distance module. The former provides the overall information of the features, which is formed by concatenating left feature maps with their corresponding right feature maps\cite{kendall2017end}.and the latter calculates the difference between the two at disparity level to provide feature similarity information\cite{chabra2019stereodrnet}.

\begin{figure*}[t]
	\centering
	\includegraphics[width=0.950\textwidth]{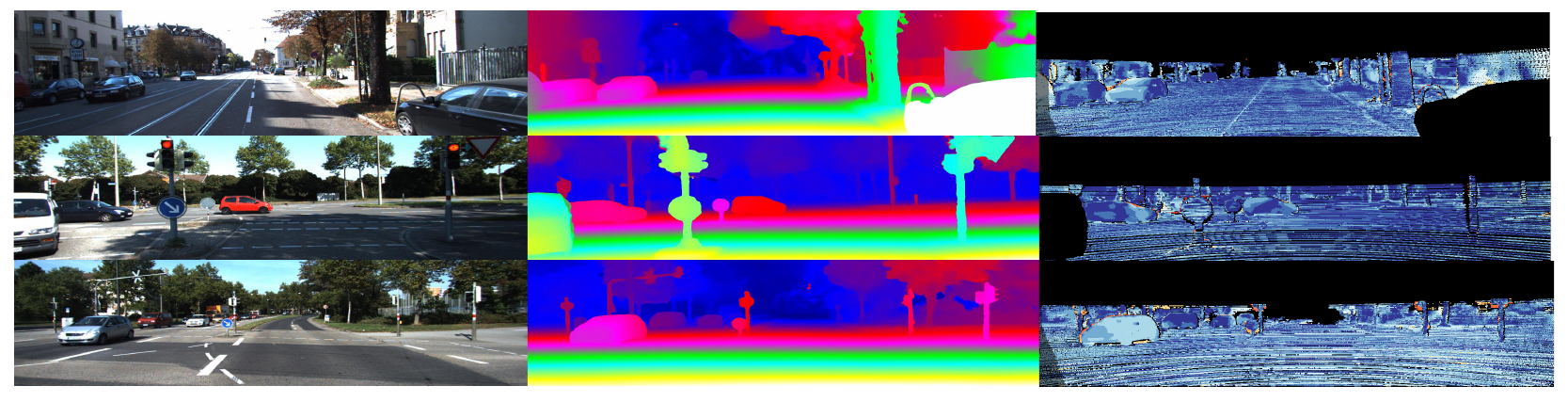} 
	\caption{Results on the KITTI 2015 test sets. From left: left stereo image, disparity map, error map.}
	\label{fig6}
\end{figure*}

\subsection{Output module and loss function}
The output module contains two stacked 3D convolution layers and the upsampling operator. The volume $c_d$ from the output module is converted into a probability volume with a softmax function $\sigma(.)$ along the disparity dimension. The predicted disparity  $\hat{d}$ calculated as follows:
\begin{equation}\label{eq6}
\hat{d} = \sum_{d=0}^{D_{max}}d\times \sigma(-c_d)
\end{equation}

The predicted disparity maps from the three output modules are denoted as $\hat{d_1}$, $\hat{d_2}$, $\hat{d_3}$, and $L_{disp}$ is as follows:
\begin{equation}\label{eq7}
L_{disp}= \sum_{i=1}^{3} \lambda_i \times Smooth_{L_{1}}\left(\hat{d_i} -d^*\right)
\end{equation}
where $\lambda$ denotes the coefficients and $d^*$ represents the ground-truth disparity map.Therefore, by combining $L_{disp}$, $L_{edge}$ and the related loss $L_{edge−disp}$, we have
\begin{equation}\label{eq8}
L_{total}  = L_{disp} + a \times L_{edge} + (1-a) \times L_{edge_disp}
\end{equation}

\section{EXPERIMENT}
In this section, we train the proposed model on the Sceneflow, Cityscapes and KITTI datasets, but evaluate it only on the Sceneflow and KITTI datasets. The disparities of the Cityscapes dataset are obtained by SGM algorithm\cite{hirschmuller2005accurate} but not the ground truth. The paper presents datasets and network implementation in Section 4.1 and Section 4.2. And illustrates the effectiveness of each module in Section 4.3 and Section 4.4. Then, the evaluation results on the different datasets are presented.

\begin{table*}[]
	\centering
	\small
	\begin{tabular}{c|cccc|ccc}
		\hline
		MODEL              & \multicolumn{4}{c|}{MODULE}                                                                                   & \multicolumn{3}{c}{RESULT}                     \\ \cline{2-8} 
		& Hourglass                         & Cost volume                   & Hard parameter sharing    & SPP          & Parameters & Sceneflow(EPE) & KITTI 2015(D1-all) \\ \hline
		PSM                  & \cite{chang2018pyramid}  &\cite{chang2018pyramid}  & -                &\cite{chang2018pyramid} & 5.27M      & 0.884         & 1.67               \\
		AGMNet            & \checkmark                       &\cite{chang2018pyramid}  & -                &\cite{chang2018pyramid} & 3.85M      & 0.801         & 1.62               \\
		AGMNet            & \checkmark                       & \checkmark                      & -                &\cite{chang2018pyramid} & 3.88M      & 0.754         & 1.56               \\
		Dedge-AGMNet & \checkmark                       & \checkmark                     & \checkmark &\cite{chang2018pyramid} & 3.88M      & 0.648         & 1.57               \\
		+Cityscapes      & \checkmark                        & \checkmark                     & \checkmark &\cite{chang2018pyramid} & 3.88M      & -                & 1.45               \\
		Dedge-AGMNet & \checkmark                       & \checkmark                      & \checkmark & \checkmark                    & 3.98M      & 0.645         & 1.54               \\
		+Cityscapes      & \checkmark                        & \checkmark                      & \checkmark & \checkmark                    & 3.98M      & -                 & 1.38               \\ \hline
	\end{tabular}
	\caption{Ablation study on the Sceneflow test set and the KITTI 2015 validation set. The symbol $'\checkmark'$ denotes the module we proposed. '+Cityscapes' denotes that the network pre-trains on the hybrid dataset, which contains Sceneflow and Cityscapes dataset.}
	\label{tab1}
\end{table*}

\begin{figure*}[t]
	\centering
	\includegraphics[width=1\textwidth]{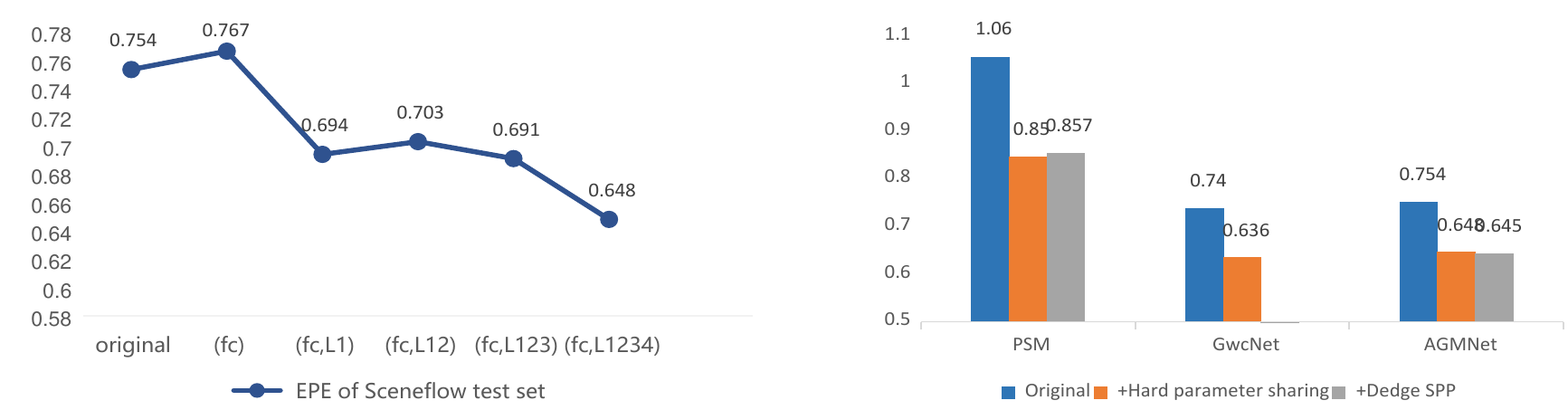} 
	\caption{\textbf{The graph}: the relationship between the performance and the shared layers. 'fc' denotes first\_conv. \textbf{The histogram}: Embedding the Depth edge auxiliary subnetwork into PSMNet\cite{chang2018pyramid}, GwcNet\cite{guo2019group-wise} and our network.The blue columns are the original results, the orange columns show the results that sharing parameters in feature extraction module. And gray columns mean that adding the hard parameter sharing mechanism and Dedge-SPP together.}
	\label{fig7}
\end{figure*}

\subsection{Datasets and evaluation metric}
\subsubsection{Stereo dataset}
\noindent \textbf{Sceneflow} is a large scale synthetic dataset containing three subsets(Flyingthings3D, Driving and Monkaa ), which provides approximately 35000 training and 4000 testing stereo image pairs of size $960\times 540$. It consists of left and right images, complete ground-truth disparity maps and segmentation images. This paper adopts end-point-error (EPE) as the evaluation metric.

\noindent \textbf{Cityscapes} is an urban scene-understanding dataset. This dataset provides 3475 rectified stereo pairs, fined annotated segmentation maps and corresponding disparity maps precomputed by SGM.

\noindent \textbf{KITTI 2012 and KITTI 2015} are both driving scene datasets. KITTI 2012 provides 194 training and 195 testing image pairs, while KITTI 2015 contains 200 training and 200 testing image pairs. With a size of $1240\times 376$, both datasets provide sparse disparity maps. Twenty image pairs have remained as the validation set. The main evaluation metric for KITTI 2015 is D1-all error, which computes the percentage of pixels for which the estimation error is $\geq3px$ or $\geq5\%$ from the ground-truth disparity. The main evaluation criterion for KITTI 2012 is Out-Noc, which computes the percentage of pixels for which the estimation error is $\geq3px$ for all non-occluded pixels.

\subsubsection{Depth edge dataset}
\noindent \textbf{Sceneflow \& Cityscapes \& KITTI 2015} According to the method proposed in Section 3.2.3, this paper generates the ground-truth of depth edges for their corresponding dataset, respectively.

\subsection{Network implementation}
The Dedge-AGMNet architecture is implemented with PyTorch. All the models are trained using the Adam optimizer($\beta_1=0.09\ \beta_2=0.999$). We use 4 Nvidia TITAN XP GPUs when training the models, and the batchsize is fixed to 8. The images are randomly Cropped to $512\times256$. The coefficients of disparity outputs are set as follows:$\lambda_1 = 0.5, \lambda_2 = 0.7, \lambda_3 = 1.0$. In $L_{edge−disp}$, $\gamma = 0.5$.

The training process of our network contains two steps. For the first step, we pre-train Dedge-AGMNet only on the Sceneflow dataset. The initial learning rate is set to 0.001, then down-scaled by 2 every 2 epochs from epoch 10 to 16. The maximum disparity ($D_{max}$) is set to 192. Besides, we fine-tune the pre-trained model with stepped learning rates of 0.001 for 300 epochs on KITTI 2012/2015. Furthermore, we extend the training to 70 epochs on Sceneflow to get the final results.

For the second step, we combine Sceneflow and Cityscapes as the pre-trained dataset. And employ the same training strategy to obtain the compared result. Finally, our network is fine-tuned with learning rates of 0.001 for 600 epochs and 0.0001 for another 400 epochs to capture the final results.

\begin{table}
	\normalsize 
	\setlength{\tabcolsep}{4mm}{
	\begin{tabular}{c|ccccc}
	\hline
	a      & 0    & 0.2  & 0.5  & 0.8  & 1    \\ \hline
	D1-all & \textbf{1.38} & 1.42 & 1.48 & 1.45 & 1.47 \\
	EPE    & \textbf{0.62} & 0.64 & 0.65 & 0.65 & 0.65 \\ \hline
	\end{tabular}}
	\caption{Control experiment for the weight a of loss function. We computed the D1-all and the EPE on the KITTI 2015 validation set.}
	\label{tab2}
\end{table}

\subsection{Effectiveness of depth edge network}
As shown in the graph of Fig.\ref{fig7}, with more and deeper shared layers in the feature extraction, the EPE decreases significantly on Sceneflow. To prove the effectiveness and generalization of depth edge auxiliary task, this paper embeds the depth edge subnetwork into PSMNet\cite{chang2018pyramid},GwcNet \cite{guo2019group-wise} and our AGMNet. As shown in the histogram, utilizing the hard parameter sharing mechanism, the depth edge subnetwork could optimize the feature extraction module. The EPE of Sceneflow is reduced about $15\% \sim 20\%$. It is worth emphasizing that this module does not add any parameters and computational cost any more. On Sceneflow dataset, the Dedge-SPP module does not improve the accuracy of the disparity map remarkably.

As shown in Table \ref{tab1}, Dedge-SPP plays a significant role to improve the accuracy on KITTI 2015. Without estimating the depth edge in the inference process, Dedge-SPP only increases few parameters(0.1M) to obtain the depth edge features. Besides, while adding Cityscapes dataset, the 3-px error is reduced clearly on KITTI validation set. This dataset guides the auxiliary subnetwork to learn more accurate features of depth edge on city scenes.

In the pre-training process, we fixed the hyper-parameter $a = 0.5$ and select the optimal setting for the weight a on KITTI 2015. As shown in Table \ref{tab2}, when $a = 0$, the validation set could obtain the best performance. It demonstrates that, for the fine-tuned network, smoothen the disparity map play a more important role than learning the depth edge feature on KITTI 2015.

\begin{table}
	\normalsize 
	\begin{tabular}{c|c|cccc|c}
		\hline
		Model & Hourglass & \multicolumn{4}{c|}{Dilation rate} & Sceneflow(EPE) \\ \hline
		PSMNet & \cite{chang2018pyramid} & \multicolumn{4}{c|}{-} & 0.889 \\
		AGMNet & \checkmark & 1 & 4 & 8 & - & 0.836 \\
		AGMNet & \checkmark & 1 & 2 & 3 & 4 & 0.823 \\
		AGMNet & \checkmark & 1 & 2 & 4 & 8 & 0.821 \\
		AGMNet & \checkmark & 1 & 4 & 8 & 16 & \textbf{0.801} \\
		AGMNet & \checkmark & 1 & 4 & 16 & 32 & 0.842 \\ \hline
	\end{tabular}
	\caption{The AGMNet network is deﬁned as the version that only replaces the hourglass structure. All the models are trained with the same learning strategy.}
	\label{tab3}
\end{table}

\subsection{Best setting of AGM module}
The experimental results in Table \ref{tab3} show that when the dilation rate is set to an appropriate range, the parallel structure with four granular convolutions can achieve better results than three. We conclude that the AGM module with dilation rates of 1, 4, 8, and 16 provides optimal performance. All the AGM-base networks outperform PSMNet. Under the best AGM module settings, the EPE is reduced by 9.3\% on Sceneflow dataset.

\begin{table*}[h]
	\centering
	\setlength{\tabcolsep}{4mm}{
		\begin{tabular}{c|cc|cc|cc|cc|cc}
			\hline
			mthod & \multicolumn{2}{l|}{$>2px(\%)$} & \multicolumn{2}{l|}{$>3px(\%)$} & \multicolumn{2}{l|}{$>4px(\%$)} & \multicolumn{2}{l}{$>5px(\%)$}& \multicolumn{2}{l}{Mean Error}  \\ \hline
			                                                                & Noc           & All            & Noc          & All            & Noc           & All           & Noc       & All      & Noc     & All   \\ \cline{2-9}
			GC-Net\cite{kendall2017end}                    & 2.71          & 3.46         & 1.77          & 2.30          & 1.36          & 1.77         & 1.12       & 1.46   &0.6       &0.7   \\
			SegStereo\cite{yang2018segstereo}         & 2.66          & 3.19         & 1.68          & 2.03         & 1.25          & 1.52         & 1.00       & 1.21    &0.5       &0.6   \\
			EdgeStereo\cite{song2018edgestereo}     &-                 &-               &1.73           &2.18          &1.30            &1.64         &1.04        &1.32     &-           &-      \\
			PSMNet\cite{chang2018pyramid}              & 2.44          & 3.01         & 1.49          & 1.89         & 1.12          & 1.42         & 0.90       & 1.15     &0.5       &0.6   \\
			EdgeStereo-v2\cite{song2019edgestereo} & 2.32          & 2.88         & 1.46          & 1.83         & 1.07          & 1.34         & 0.83       & 1.04    &\textbf{0.4}      &\textbf{0.5 }  \\ \hline
			Dedge-AGMNet      & \textbf{2.02}          & \textbf{2.56}         & \textbf{1.26}          & \textbf{1.64 }        & \textbf{0.95 }         &
			\textbf{1.24}         & \textbf{0.77 }         & \textbf{1.01 }       &\textbf{0.4}      &\textbf{0.5 }    \\ \hline
	\end{tabular}}
	\caption{Comparison with the top publiced methods on the KITTI stereo 2012 test set.}
	\label{tab4}
\end{table*}

\begin{table*}[h]
	\centering
	\setlength{\tabcolsep}{5mm}{
		\begin{tabular}{c|ccc|ccc|c}
			\hline
																		   & \multicolumn{3}{c|}{All(\%)} & \multicolumn{3}{c|}{Non-Occluded(\%)} &  Runtime                            \\ \cline{2-8}
			Method                                                   & D1-bg  & D1-fg     & D1-all  & D1-bg    & D1-fg     & D1-all    & (s)      \\ \hline
			GC-Net\cite{kendall2017end}                  & 2.21     & 6.16       & 2.87    & 2.02       & 5.58      & 2.61     & 0.9s    \\ 
			PSMNet\cite{chang2018pyramid}            & 1.86     & 4.62      & 2.32    & 1.71        & 4.31       & 2.14     & 0.41s   \\ 
			SegStereo\cite{yang2018segstereo}        & 1.88     & 4.07      & 2.25    & 1.76       & 3.70       & 2.08     & 0.6s      \\ 
			EdgeStereo\cite{song2018edgestereo}    & 1.87     & 3.61      & 2.16     & 1.72       & 3.41        & 3.00    & 0.7s     \\
		EdgeStereo-v2\cite{song2019edgestereo}   &1.84      & 3.30     & 2.08    &1.69         & 2.94       & 1.89    & \textbf{0.32s }   \\ \hline
			AGMNet                                                  &1.66      &4.30       &2.10     &1.53        &3.89         &1.92     &0.84s                   \\ 
			Dedge-AGMNet                                      & \textbf{1.54}    & \textbf{3.37} & \textbf{1.85}  & \textbf{1.41}     & \textbf{2.98}    & \textbf{1.67}     & 0.9s    \\ \hline
	\end{tabular}}
	\caption{Comparison with the top publiced methods on the KITTI stereo 2015 test set.}
	\label{tab5}
\end{table*}

\begin{table}
	\setlength{\tabcolsep}{3.2mm}{
	\begin{tabular}{cc|cc|cc}
	\hline
	Mod. & EPE & Mod. & EPE & Mod. & EPE \\ \hline
	GC-Net & 2.51 & SegStereo & 1.45 & PSMNet & 1.09 \\
	CSPN & 0.78 & AMNet32 & 0.74 & ours & \textbf{0.520 }\\ \hline
	\end{tabular}}
	\caption{Comparison with the top publiced methods on the Sceneflow test set.}
	\label{tab6}
\end{table}

\begin{figure}[t]
	\centering
	\includegraphics[width=1\columnwidth]{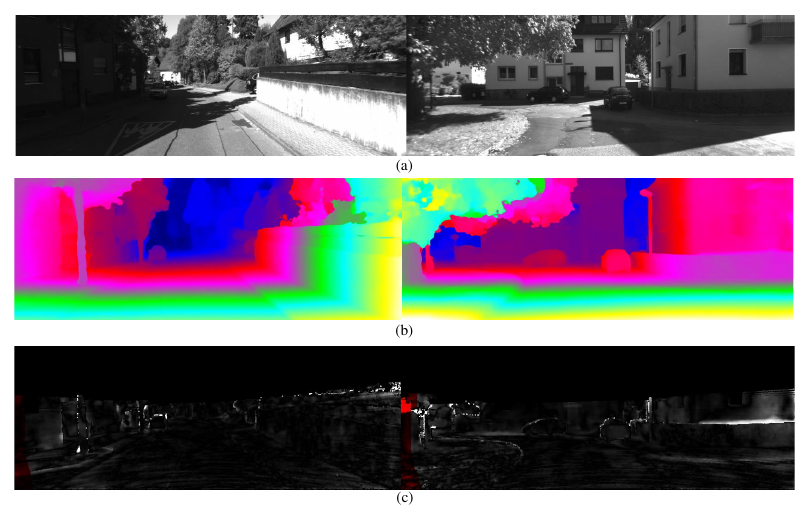} 
	\caption{Results on the KITTI 2012 test sets. (a) denotes left stereo image, (b) denotes disparity map and (c) presents the error map.}
	\label{fig8}
\end{figure}
\subsection{Result}
As shown in Table \ref{tab1}, the result shows that the module we proposed has a certain effect to promote the network. Compared to PSMNet\cite{chang2018pyramid}, our proposed network has a 27.0\% reduction on Sceneflow and 17.4\% on the KITTI 2015 validation dataset.

\textbf{Sceneflow}: We compared the performance of PSMNet with other state-of-the-art methods, such CSPN\cite{cheng2018learning}, AMNet32\cite{du2019amnet}. As shown in Table 6, Dedge-AGMNet ranks first compared to other published papers, which shows the effectiveness of the depth edge auxiliary task thoroughly.

\textbf{KITTI 2012 and 2015}: Our approach achieves state-of-the-art performances on KITTI 2012 and KITTI 2015 benchmark datasets.Utilizing the best hyper-parameter setting selected in the experiment, we train our model for 1000 epochs on KITTI 2015. Then estimate the disparity maps for the 200 testing images. According to the online leaderboard, as shown in Table \ref{tab5}, the D1-all for the Dedge-AGMNet is 1.85\%, which ranks in the \textbf{fourth} place. Similarly, we calculate the disparity for the KITTI 2012 test set. As shown in Table \ref{tab4}, the result ranks \textbf{fourth}, too. 

Fig.\ref{fig6} and Fig.\ref{fig8} give qualitative results on the KITTI 2012 and 2015 test sets, which demonstrates that our network produces high-quality results in ill-regions.

\section{CONCLUSION}
In this paper, we propose the Dedge-AGMNet, a stereo matching network optimized by depth edge. This paper expounds on the superiority of the auxiliary depth edge task and generates the depth edge ground-truth innovatively. Dedge-AGMNet contains two main modules: Depth edge subnetwork and AGM module. We utilize the hard parameter sharing mechanism to joint optimize the feature extraction module. And design Dedge-SPP to fuse the depth edge features. The proposed AGM module provides multi-scale context information while consuming fewer computational resources. The ablation study demonstrates the effectiveness of the above modules. In our experiment, Dedge-AGMNet achieves state-of-the-art performances and outperforms other multi-task learning models. The proposed network ranks in the first place on Sceneflow and fourth place on both KITTI 2012 and 2015. In the future, we plan to apply the depth edge auxiliary task on a real-time stereo matching network.

\ack This work is supported by Science and Technology Planning Project of Shenzhen(JCYJ20180503182133411).

\bibliography{ecai}

\begin{thebibliography}{10}

\bibitem{chabra2019stereodrnet}
Rohan Chabra, Julian Straub, Christopher Sweeney, Richard Newcombe, and Henry
  Fuchs, `Stereodrnet: Dilated residual stereonet', in {\em Proceedings of the
  IEEE Conference on Computer Vision and Pattern Recognition}, pp.
  11786--11795, (2019).

\bibitem{chang2018pyramid}
Jia-Ren Chang and Yong-Sheng Chen, `Pyramid stereo matching network', in {\em
  Proceedings of the IEEE Conference on Computer Vision and Pattern
  Recognition}, pp. 5410--5418, (2018).

\bibitem{chen2017rethinking}
Liang-Chieh Chen, George Papandreou, Florian Schroff, and Hartwig Adam,
  `Rethinking atrous convolution for semantic image segmentation', {\em arXiv
  preprint arXiv:1706.05587}, (2017).

\bibitem{cheng2018learning}
Xinjing Cheng, Peng Wang, and Ruigang Yang, `Learning depth with convolutional
  spatial propagation network', {\em arXiv preprint arXiv:1810.02695}, (2018).

\bibitem{du2019amnet}
Xianzhi Du, Mostafa El-Khamy, and Jungwon Lee, `Amnet: Deep atrous multiscale
  stereo disparity estimation networks', {\em arXiv preprint arXiv:1904.09099},
  (2019).

\bibitem{gao2019res2net}
Shang-Hua Gao, Ming-Ming Cheng, Kai Zhao, Xin-Yu Zhang, Ming-Hsuan Yang, and
  Philip Torr, `Res2net: A new multi-scale backbone architecture', {\em arXiv
  preprint arXiv:1904.01169}, (2019).

\bibitem{guo2019group-wise}
Xiaoyang Guo, Kai Yang, Wukui Yang, Xiaogang Wang, and Hongsheng Li,
  `Group-wise correlation stereo network',  3273--3282, (2019).

\bibitem{hirschmuller2005accurate}
Heiko Hirschmuller, `Accurate and efficient stereo processing by semi-global
  matching and mutual information', in {\em 2005 IEEE Computer Society
  Conference on Computer Vision and Pattern Recognition (CVPR'05)}, volume~2,
  pp. 807--814. IEEE, (2005).

\bibitem{huang2017densely}
Gao Huang, Zhuang Liu, Laurens Van Der~Maaten, and Kilian~Q Weinberger,
  `Densely connected convolutional networks', in {\em Proceedings of the IEEE
  conference on computer vision and pattern recognition}, pp. 4700--4708,
  (2017).

\bibitem{kendall2017end}
Alex Kendall, Hayk Martirosyan, Saumitro Dasgupta, Peter Henry, Ryan Kennedy,
  Abraham Bachrach, and Adam Bry, `End-to-end learning of geometry and context
  for deep stereo regression', in {\em Proceedings of the IEEE International
  Conference on Computer Vision}, pp. 66--75, (2017).

\bibitem{long2015fully}
Jonathan Long, Evan Shelhamer, and Trevor Darrell, `Fully convolutional
  networks for semantic segmentation', in {\em Proceedings of the IEEE
  conference on computer vision and pattern recognition}, pp. 3431--3440,
  (2015).

\bibitem{luo2016efficient}
Wenjie Luo, Alexander~G Schwing, and Raquel Urtasun, `Efficient deep learning
  for stereo matching', in {\em Proceedings of the IEEE Conference on Computer
  Vision and Pattern Recognition}, pp. 5695--5703, (2016).

\bibitem{mayer2016large}
Nikolaus Mayer, Eddy Ilg, Philip Hausser, Philipp Fischer, Daniel Cremers,
  Alexey Dosovitskiy, and Thomas Brox, `A large dataset to train convolutional
  networks for disparity, optical flow, and scene flow estimation', in {\em
  Proceedings of the IEEE Conference on Computer Vision and Pattern
  Recognition}, pp. 4040--4048, (2016).

\bibitem{schops2017multi}
Thomas Schops, Johannes~L Schonberger, Silvano Galliani, Torsten Sattler,
  Konrad Schindler, Marc Pollefeys, and Andreas Geiger, `A multi-view stereo
  benchmark with high-resolution images and multi-camera videos', in {\em
  Proceedings of the IEEE Conference on Computer Vision and Pattern
  Recognition}, pp. 3260--3269, (2017).

\bibitem{seki2017sgm}
Akihito Seki and Marc Pollefeys, `Sgm-nets: Semi-global matching with neural
  networks', in {\em Proceedings of the IEEE Conference on Computer Vision and
  Pattern Recognition}, pp. 231--240, (2017).

\bibitem{song2019edgestereo}
Xiao Song, Xu~Zhao, Liangji Fang, and Hanwen Hu, `Edgestereo: An effective
  multi-task learning network for stereo matching and edge detection', {\em
  arXiv preprint arXiv:1903.01700}, (2019).

\bibitem{song2018edgestereo}
Xiao Song, Xu~Zhao, Hanwen Hu, and Liangji Fang, `Edgestereo: A context
  integrated residual pyramid network for stereo matching', in {\em Asian
  Conference on Computer Vision}, pp. 20--35. Springer, (2018).

\bibitem{xie2015holistically}
Saining Xie and Zhuowen Tu, `Holistically-nested edge detection', in {\em
  Proceedings of the IEEE international conference on computer vision}, pp.
  1395--1403, (2015).

\bibitem{yang2018segstereo}
Guorun Yang, Hengshuang Zhao, Jianping Shi, Zhidong Deng, and Jiaya Jia,
  `Segstereo: Exploiting semantic information for disparity estimation', in
  {\em Proceedings of the European Conference on Computer Vision (ECCV)}, pp.
  636--651, (2018).

\bibitem{yang2018denseaspp}
Maoke Yang, Kun Yu, Chi Zhang, Zhiwei Li, and Kuiyuan Yang, `Denseaspp for
  semantic segmentation in street scenes', in {\em Proceedings of the IEEE
  Conference on Computer Vision and Pattern Recognition}, pp. 3684--3692,
  (2018).

\bibitem{yu2017casenet}
Zhiding Yu, Chen Feng, Ming-Yu Liu, and Srikumar Ramalingam, `Casenet: Deep
  category-aware semantic edge detection', in {\em Proceedings of the IEEE
  Conference on Computer Vision and Pattern Recognition}, pp. 5964--5973,
  (2017).

\bibitem{zbontar2016stereo}
Jure Zbontar, Yann LeCun, et~al., `Stereo matching by training a convolutional
  neural network to compare image patches.', {\em Journal of Machine Learning
  Research}, {\bf 17}(1-32), ~2, (2016).

\end{thebibliography}
\end{document}